\pdfoutput=1

\documentclass[11pt]{article}

\usepackage{acl}

\usepackage{times}
\usepackage{latexsym}

\usepackage[T1]{fontenc}

\usepackage[utf8]{inputenc}

\usepackage{microtype}
\usepackage{xcolor}
\usepackage{graphicx}
\usepackage{cleveref}
\usepackage{caption}
\usepackage{subcaption}

\title{Better Retrieval May Not Lead to Better Question Answering}

\author{Zhengzhong Liang$^{\clubsuit}$, Tushar Khot$^{\spadesuit}$, Steven Bethard$^{\diamondsuit}$, \\
{\bf Mihai Surdeanu$^{\clubsuit}$}, {\bf Ashish Sabharwal$^{\spadesuit}$} \\
$^{\clubsuit}$Computer Science Department, The University of Arizona, Tucson, AZ \\
$^{\diamondsuit}$School of Information, The University of Arizona, Tucson, AZ \\
$^{\spadesuit}$Allen Institute for Artificial Intelligence, Seattle, WA \\
\texttt{\{zhengzhongliang, bethard, msurdeanu\}@email.arizona.edu} \\
\texttt{\{tushark, ashishs\}@allenai.org}} 

\begin{document}
\maketitle
\begin{abstract} 
Considerable progress has been made recently in open-domain question answering (QA) problems, which require Information Retrieval (IR) and Reading Comprehension (RC). A popular approach to improve the system's performance is to improve the quality of the retrieved context from the IR stage. In this work we show that for StrategyQA, a challenging open-domain QA dataset that requires multi-hop reasoning, this common approach is surprisingly ineffective---improving the quality of the retrieved context hardly improves the system's performance. We further analyze the system's behavior to identify potential reasons.
\end{abstract}

\section{Introduction}
Open-domain Question Answering (QA) refers to the problem of answering questions \emph{without} any provided context~\cite{voorhees1999trec}. Systems are often expected to search for relevant context in a large-scale corpus and use this context to answer the questions, e.g., using Wikipedia to answer trivia questions~\cite{chen-etal-2017-reading, yang-etal-2018-hotpotqa, kwiatkowski-etal-2019-natural}. Due to the nature of the task, solutions to open-domain QA~\cite{chen-etal-2017-reading, nie-etal-2019-revealing, qi-etal-2021-answering} often break it down into two sub-problems: evidence retrieval (to retrieve the relevant context) and QA (to answer the question given the relevant context).

Since each of these sub-problems can be challenging, prior work has often focused on one sub-problem at a time---improving retrieval models using gold context supervision~\cite{nie-etal-2019-revealing, karpukhin-etal-2020-dense, mao-etal-2021-generation, xiong2020answering, zhao-etal-2021-multi-step} or improving QA given relevant context~\cite{qiu-etal-2019-dynamically, jiang-bansal-2019-self, zaheer2020big, groeneveld-etal-2020-simple}. While this has been a successful strategy \cite{nie-etal-2019-revealing, karpukhin-etal-2020-dense, mao-etal-2021-generation}, it assumes that improved retrieval will translate to improvement in down-stream QA performance. In this work, we surprisingly show that this is not always the case.

Specifically, we show that on the open-domain multi-hop QA dataset  StrategyQA~\cite{geva-etal-2021-aristotle}, improvement in retrieval quality has marginal to no effect on the accuracy of a state-of-the-art QA model~\cite{khashabi-etal-2020-unifiedqa}. We contrast this behavior with another popular multi-hop QA dataset, HotpotQA~\cite{yang-etal-2018-hotpotqa}, where gradual improvement in retrieval quality does result in a corresponding improvement in QA scores. Further study reveals that this surprising behavior on StrategyQA could be attributed to the Boolean nature of the questions and the models not being able to leverage even the gold evidence very well.

We make two key contributions: 
(1) We show that purely focusing on retrieval may not always translate into improved QA performance; 
(2) We analyze the StrategyQA dataset to identify possible reasons for this behavior.
Our work suggests that improving retrieval quality is a surprisingly unproductive path (using current models) for the open-domain QA problems with Boolean questions which require diverse reasoning skills and commonsense knowledge implicit in the question.\footnote{\raggedright The code and data can be found at: \url{https://github.com/zhengzhongliang/BetterRetrievalNotBetterQA}}

 \begin{figure*}
     \centering
     \begin{subfigure}[b]{0.29\textwidth}
         \centering
         \includegraphics[width=\textwidth]{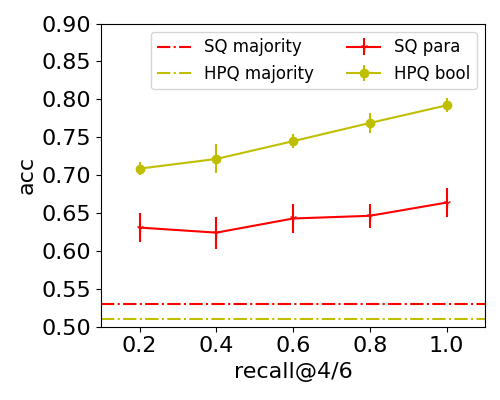}
         \caption{Test acc.~of StrategyQA and HotpotQA-Boolean}
         \label{fig:acc_sq_hq_bool}
     \end{subfigure}
     \hfill
     \begin{subfigure}[b]{0.29\textwidth}
         \centering
         \includegraphics[width=\textwidth]{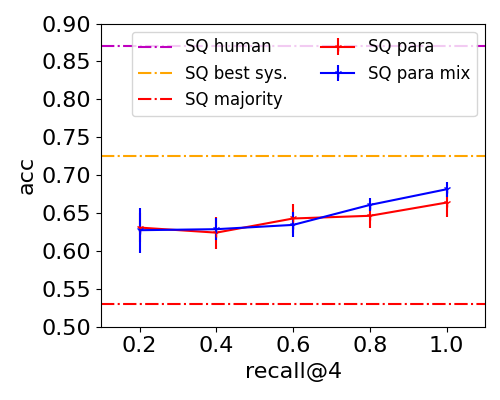}
         \caption{Test acc.~of StrategyQA with mixed annotations}
         \label{fig:acc_sq_para_mix}
     \end{subfigure}
     \hfill
     \begin{subfigure}[b]{0.29\textwidth}
         \centering
         \includegraphics[width=\textwidth]{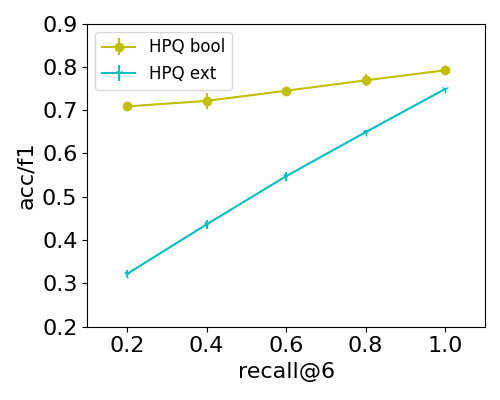}
         \caption{Test acc.~of HotpotQA-Boolean and HotpotQA-extractive}
         \label{fig:acc_hq_bool_extractive}
     \end{subfigure}
\caption{The model's QA performance on StrategyQA, HotpotQA-Boolean and HotpotQA-extractive. The bar shows standard deviation across 9 experiments 3 data seeds $\times$ 3 model seeds). For the Boolean datasets, we also show the majority voting baselines. The human performance in \cref{fig:acc_sq_para_mix} is discussed in \cref{sec:app_manual_analysis}. }
\label{fig:main_result}
\end{figure*}

\section{Experimental Setup}
\label{sec:exp_design}

\subsection{Task Formulation}
We formulate the task of open-domain QA as a two step process: (1) retrieve relevant documents, and (2) answer the question given the retrieved documents. Given a question $Q$, our system retrieves the relevant documents $C$ and predicts the answer $A$ using a reading comprehension (RC) model $M$, i.e.,  $A=M(Q, C)$. We assume that the retrieval approach will retrieve $k$ paragraphs/documents $C=\{P_1, P_2, \dots, P_k\}$ and optimize for the standard recall$@k$ metric, denoted $m$. Our goal is to assess the impact of improvement in this metric $m$ on the final QA accuracy of the system. We do so by simulating retrieval approaches with varying values of $m$ as described in \cref{subsec:poison}.



\subsection{Datasets}
\paragraph{StrategyQA} This is our primary dataset, containing multi-hop open-domain Boolean questions that require \emph{implicit decomposition}, i.e., the strategy to answer the question is not explicitly described in the question itself. For examples, "Did Aristotle use a laptop?" requires comparison of their time periods. The dataset has about 2K questions with about 2-4  paragraphs annotated as gold evidence.\footnote{For an example of StrategyQA/HotpotQA, please see \cref{tab:examples_hotpotqa_strategyqa} in the Appendix. }



\paragraph{HotpotQA} We use HotpotQA~\cite{yang-etal-2018-hotpotqa}, also a multi-hop QA dataset, to contrast against the behavior of models on StrategyQA. HotpotQA contains questions such as "Who is older, Mark Aguirre or Adrian Dantley?" where the decompositions are more explicit.\footnote{For the possible impact of the implicit decomposition, please see \cref{sec:app_manual_analysis}.} We use both the extractive questions (referred to as HPQ ext) and Boolean questions (referred to as HPQ bool).



\subsection{Model}
\label{subsec:poison}

\paragraph{Retrieval Output:} To evaluate the impact of retrieval quality on QA performance, we construct contexts with varying recall values.
Specifically, we start from the gold supporting paragraphs and randomly replace (with probability $m$) a gold paragraph with non-gold distractors from the top retrieved paragraphs. We add more distractor paragraphs, if needed, to have $k$ paragraphs in the context.\footnote{$k$ is set to 4 StrategyQA and 6 for HotpotQA to ensure similar context lengths for both datasets. See \cref{sec:appendix_poisoning} for more details.} 

To simulate the quality of the retrieved paragraphs in a real application as much as possible, we use deliberately designed retrieval and reranking methods to find the candidate distractors. For HotpotQA, we use the reranked paragraphs retrieved by the pipeline proposed in \citet{nie-etal-2019-revealing}. For StrategyQA, we use a pipeline of ``question decomposition -> retrieval -> reranking'' that eventually yields 0.457 Recall$@$10 on the test partition.\footnote{See \cref{sec:appendix_poisoning} for more details.} In comparison, the retrieval baseline in the original StrategyQA paper \cite{geva-etal-2021-aristotle} only yields a Recall$@$10 of 0.221 on the same partition.

\paragraph{QA Model:}
Our system uses UnifiedQA-large \cite{khashabi-etal-2020-unifiedqa}, which is based on T5-large~\cite{raffel2020exploring}, as the RC model for both HotpotQA and StrategyQA. UnifiedQA has been shown to outperform multiple state-of-the-art models on a variety of datasets including commonsense QA datasets (e.g., the PhysicalIQA dataset~\cite{Bisk2020}), a skill needed for StrategyQA. We use a similar fine-tuning regimen as proposed by \citet{geva-etal-2021-aristotle} and achieve an accuracy of 0.725, comparable to the published RoBERTa model's accuracy of 0.723.\footnote{Both models are trained and evaluated on gold context.} See \cref{app:uqa_details} for details.

\subsection{Experimental Details}
We create a train/dev/test split with 1855/209/229 examples for StrategyQA, 1855/209/458 examples for HotpotQA-Boolean and 1855/209/6947 examples for HotpotQA-extractive.\footnote{We create our own dev/test split since the test set answers are not known. Additionally this also ensures that we have the same number of train and dev examples in each dataset.} We consider five difference recall values: $m \in \{0.2, 0.4, 0.6, 0.8, 1.0\}$ and generate contexts for each recall value. Due to the sampling in our dataset and context construction, we experiment with 3 seeds for the sampling process in addition to 3 seeds for the QA model training. As a result, we have a total of 9 experiments for each value of p and report the mean and standard deviation of the QA model accuracy on the test set.

\begin{figure}
     \centering
     \begin{subfigure}[b]{0.23\textwidth}
         \centering
         \includegraphics[width=\textwidth]{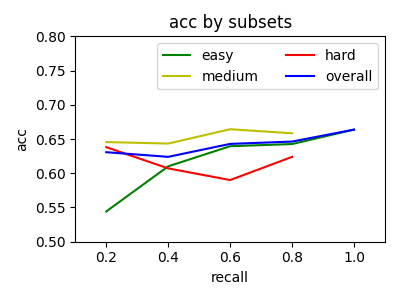}
         \caption{Acc.~by subsets of SQ}
         \label{fig:sq_acc_by_subsets}
     \end{subfigure}
     \begin{subfigure}[b]{0.23\textwidth}
         \centering
         \includegraphics[width=\textwidth]{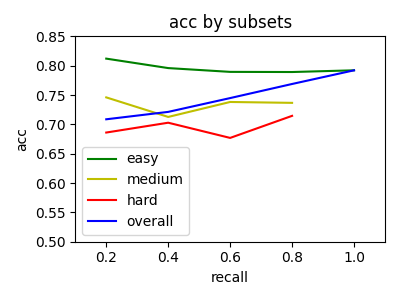}
         \caption{Acc.~by subsets of HPQ}
         \label{fig:hq_acc_by_subsets}
     \end{subfigure}
     
     \begin{subfigure}[b]{0.23\textwidth}
         \centering
         \includegraphics[width=\textwidth]{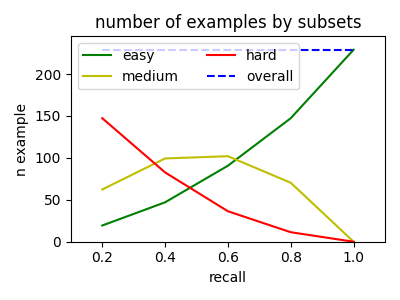}
         \caption{$\#$N examples of SQ}
         \label{fig:sq_n_example_by_subsets}
     \end{subfigure}
     \begin{subfigure}[b]{0.23\textwidth}
         \centering
         \includegraphics[width=\textwidth]{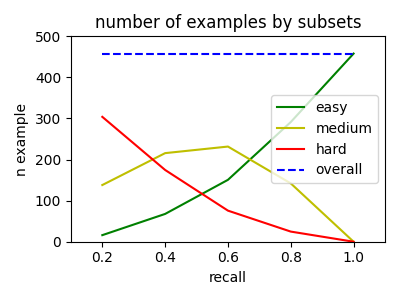}
         \caption{$\#$N examples of HPQ}
         \label{fig:hq_n_example_by_subsets}
     \end{subfigure}
\caption{The test accuracy of different subsets of StrategyQA and HotpotQA-Boolean. \cref{fig:sq_n_example_by_subsets} and \cref{fig:hq_n_example_by_subsets} shows the number of examples of different subsets of StrategyQA and HotpotQA-Boolean.}
\label{fig:acc_by_subsets}
\end{figure}
\section{Results}
We compare the accuracy of the UnifiedQA model on context with varying recall values in \cref{fig:main_result}. 

\paragraph{Finding 1: Evidence quality minimally impacts StrategyQA}
As shown by the \emph{SQ para} line in \cref{fig:acc_sq_hq_bool}, increasing the recall from 0.2 to 1.0 has negligible effect on the QA model's accuracy. 
 Even after a 5x increase in recall to achieve an almost perfect retrieval system (100\% recall), the QA model score only goes from 63\% to 66\%. Additionally as shown by the standard deviation, the gains are not even statistically significant. 

\paragraph{Finding 2: Improved retrieval does help other Boolean multi-hop datasets} Given these results, a natural conclusion would be to attribute this behavior to either multi-hop or Boolean nature of the questions. Hence, we compare the performance of our QA model on the similarly poisoned contexts from HotpotQA Boolean questions, denoted \emph{HPQ bool} in \cref{fig:acc_sq_hq_bool}). In this case, the model's QA accuracy increases almost linearly from 70.9\% to 79.8\% as the recall increases from 0.2 to 1.0, and the standard deviations are small, such that each increase is significant.

\paragraph{Finding 3: Increasing evidence supervision has minimal impact on StrategyQA} Since there are only 1,855 training examples in the training set, we further investigate whether increasing the number of training examples could potentially help. Instead of randomly selecting one evidence annotation per question (c.f. \cref{sec:appendix_poisoning_strategyqa}), we use all evidence annotations for each question, resulting in about 5,500 training examples. This setting is denoted \emph{SQ para mix} in \cref{fig:acc_sq_para_mix}. While this increased training data has a minor impact when the recall $\geq 0.8$, the gains are not significant. Note that developing retrieval systems with such high recall$@4$ would be a very challenging task, especially in StrategyQA where the questions require multiple documents. Our best performing system for StrategyQA by far is to use the gold paragraphs from all annotators for each question without distractors, achieving 72.5\% accuracy. This shows that unless the retrieval is perfect, it does not drastically help with QA.

\paragraph{Finding 4: Extractive questions are more sensitive to evidence quality than Boolean questions} We next compare the model's accuracy on the poisoned HotpotQA Boolean questions to its F1 score on the poisoned HotpotQA extractive questions. As shown in \cref{fig:acc_hq_bool_extractive}, the QA performance increases almost linearly from 0.3 to 0.7 when the recall increases from 0.2 to 1.0 on HPQ-extractive subset, compared to the relatively flatter curve for Boolean questions. This indicates that it is harder for the Boolean questions to obtain QA improvement from retrieval than extractive questions.

\begin{figure*}
     \centering
     \begin{subfigure}[b]{0.24\textwidth}
         \centering
         \includegraphics[width=\textwidth]{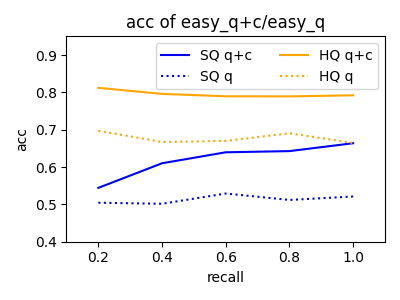}
         \caption{acc.~``q+c'' vs ``q''}
         \label{fig:easy_qc_q}
     \end{subfigure}
     \begin{subfigure}[b]{0.24\textwidth}
         \centering
         \includegraphics[width=\textwidth]{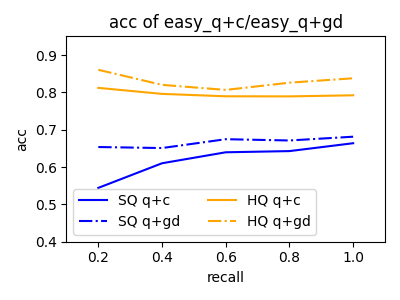}
         \caption{acc.~``q+c'' vs ``q+gd''}
         \label{fig:easy_qc_qgd}
     \end{subfigure}
     \begin{subfigure}[b]{0.24\textwidth}
         \centering
         \includegraphics[width=\textwidth]{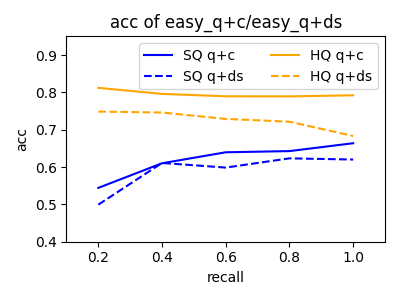}
         \caption{acc.~``q+c'' vs ``q+ds''}
         \label{fig:easy_qc_qds}
     \end{subfigure}
     \begin{subfigure}[b]{0.24\textwidth}
         \centering
         \includegraphics[width=\textwidth]{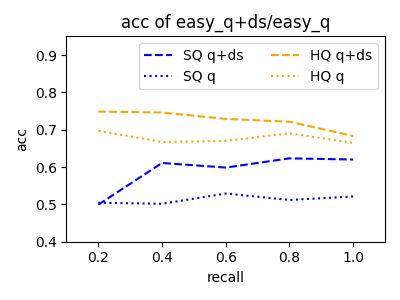}
         \caption{acc.~``q'' vs ``q+ds''}
         \label{fig:easy_q_qds}
     \end{subfigure}
\caption{Contrastive evaluation on the \textbf{easy} subset of HotpotQA-Boolean and StrategyQA. The models are trained on ``q+c'', i.e., question and the context with different poisoning levels (of all subsets), then evaluated on different inputs (but only on the easy subset). ``q'' is only question; ``q+gd'' is question and gold paragraphs (i.e., all examples having 1.0 recall and no distractors); ``q+ds'' is question and distractors; ``q+c'' is the question and the context (on this easy subset, the context ``c'' contains ``gd'' and ``ds'').}
\label{fig:sq_hq_contrastive}
\end{figure*}

\section{Explaining the QA-Retrieval Quality Independence}
\label{sec:explain_subset}
In this section we conduct experiments and manual analysis for some examples to reveal the possible reasons of the QA model's behavior on StrategyQA.

\subsection{QA Performance by Subsets}
To explain this independence of QA model's performance to retrieval quality, we analyze the impact of context on the QA model. We first evaluate the performance of the QA model on different subsets within each recall bin based on their hardness: easy (recall = 1), med (1 > recall > 0), hard (recall=0).\footnote{Note that since we are randomly sampling context, we will have some easy examples even when the aggregate recall=0.2.} One would expect models to do better on the easy subset since the context contains all the gold paragraphs. As shown in \cref{fig:acc_by_subsets}, while the accuracy on easy examples is much higher than medium and hard examples in HotpotQA, the same is not true for StrategyQA. Across all recall bins, the accuracy on easy and medium examples is roughly the same indicating that the model does not very successfully learn to exploit the gold paragraphs, even on the easy examples. Such difference largely explains the different sensitivity to evidence quality for SQ and HPQ-Boolean. For HPQ-Boolean, the number of easy examples in the test dataset increases as the recall increases (\cref{fig:hq_n_example_by_subsets}), and such a high accuracy on the easy examples causes a steeper slope. 

\subsection{Contrastive Evaluation}
We next focus on understanding: \emph{Why does the QA model not learn to exploit the gold paragraphs in the context?} We explore this question by evaluating the QA models on ablated inputs, e.g question only, question with gold paragraphs (\cref{fig:sq_hq_contrastive}).

The first natural question is: \emph{Does the QA model even use the context?} We compare the models performance on question (q) and question+context (q+c) in Fig.~\ref{fig:easy_qc_q}. On both the datasets, we can see that the q+c model  outperforms the q model indicating that the model does use the context.


We next ask the question: \emph{Does the model learn to use the gold paragraphs for QA?}. As we show in Fig.~\ref{fig:easy_qc_qgd}, the QA model for StrategyQA generally underperforms HotpotQA, even when given the gold paragraphs (gd). Hence the QA model is not able to exploit the gold paragraphs as much as HotpotQA.
It also shows that both models tend to be affected by the distractors even if the gold evidence is in the context.

Finally we evaluate the value of the distractor paragraphs (ds) on these models. As shown in Fig.~\ref{fig:easy_qc_qds}, HotpotQA has the expected behavior where the model performance drops when the context is replaced with only distractor paragraphs. However, the model for StrategyQA can answer the questions almost equally well using the distractor paragraphs showing that it does rely on these paragraphs. It is further corroborated by Fig.~\ref{fig:easy_q_qds} where the model does much better with distractor paragraphs than just the question.  

Further comparison shows that the HotpotQA model tends to use the context to answer the question when the context contains good evidence. If the context is not good, the HotpotQA model tends to use the question, instead of relying on the distractors. Surprisingly, HotpotQA model learns to use the gold paragraphs even when only 3\% of training examples (about 55 examples) have perfect evidence (recall=0.2). In contrast, the StrategyQA model relies heavily on the distractors regardless of the context quality (\cref{sec:app_ablating_context_all_subsets}) and does not learn to only use the gold paragraphs even when recall=1.0.

In summary, the StrategyQA model uses the context to answer the question, but does not learn to leverage the gold paragraphs and relies more on distractors compared to HotpotQA.

\begin{table*}
    \small
    \centering
    \begin{tabular}{p{0.5cm} | p{3.5cm} p{3cm} p{7.5cm}} \hline
      \textbf{Row} & \textbf{Q.} & \textbf{Decomp. Q.} & \textbf{Doc.}  \\ \hline \hline
      1 & Are more people today related to Genghis Khan than Julius Caesar? & How many kids did Julius Caesar have? & (\textbf{Julius Caesar-75}) Children \textcolor{blue}{Julia}, by Cornelia, born in 83 or 82 BC Caesarion, ... . \textcolor{blue}{Posthumously adopted: Gaius Julius Caesar Octavianus}, his great-nephew ... . \textcolor{blue}{Suspected Children Marcus Junius Brutus} (born 85 BC): ... \\
       & & How many kids did Genghis Khan have? & (\textbf{Genghis Khan-17}) ... \textcolor{red}{Temujin had many other children with other wives, but they were excluded from succession, only Borte's sons could be considered to be his heirs.} \\
       
      2 & Would a dog respond to bell before Grey seal? & How sensitive is a grey seal's hearing on land? & (\textbf{Pinniped-24}) ... In air, \textcolor{blue}{hearing is somewhat reduced in pinnipeds compared to many terrestrial mammals.} While they are capable of hearing a wide range of frequencies ..., their airborne \textcolor{blue}{hearing sensitivity is weaker overall.} ... \\ 
       & & How sensitive is a dog's hearing on land? & (\textbf{Dog anatomy-114}) The frequency range of dog hearing is between ..., which means that \textcolor{red}{dogs can detect sounds far beyond the upper limit of the human auditory spectrum.}  \\
       
       3 & Is average number of peas in a pod enough commas for a billion? & How many peas are in the average pod? & (\textbf{Pea-1}) The pea is most commonly the small spherical seed or the seed-pod of the pod fruit Pisum sativum. Each pod contains \textcolor{blue}{several peas}, ... \\
        & & How many commas are needed for a billion? & (\textbf{Billion-2}) \textcolor{red}{1,000,000,000}, i.e. one thousand million, or 109 (ten to the ninth power), as defined on the short scale. This is now the meaning in both British and ... \\ 
        
        4 & Would a Monoamine Oxidase candy bar cheer up a depressed friend? & Depression is caused by low levels of what chemicals? & (\textbf{Depression (mood)-13}) Alcohol can ... . \textcolor{blue}{It also lowers the level of serotonin in our brain, which could potentially lead to higher chances of a depressive mood.} \\
        & & Monoamine Oxidase has an effect on what chemicals? & (\textbf{Monoamine oxidase-7}) \textcolor{red}{Serotonin, ... and epinephrine are mainly broken down by MAO-A.} ... \\
        \hline
    \end{tabular}
    \caption{Examples from StrategyQA that are challenging for the current QA models. The Q. column shows the compositional question that the model needs to solve, and Decomp Q. shows the gold decomposed questions provided by the annotators. For each decomposed question, we provide the corresponding gold paragraph with the most relevant sentences highlighted. }
    \label{tab:strategyqa_hard_examples}
\end{table*}
 
\subsection{Are the Questions in StrategyQA Unsolvable?}
Given that (1) the StrategyQA model has a relatively low performance on the easy subset of the questions compared to the HotpotQA model (\cref{fig:acc_by_subsets}) and (2) when the recall$@$4=1.0 the StrategyQA model only reaches an accuracy of 0.68 (\cref{fig:acc_sq_para_mix}), one might argue that the questions in StrategyQA are unsolvable.
However, human annotators were able to answer the questions with an accuracy of about 0.87 \cite{geva-etal-2021-aristotle}, indicating most questions in StrategyQA are solvable by human. This accuracy is computed on 100 randomly sampled questions. The human participants can use Wikipedia articles (but the gold paragraphs are not provided) and can optionally choose to check the gold decomposed questions (however, the usage of gold decomposed questions turned out to be relatively rare, only 14\%). The study also shows that world knowledge is widely used due to the relatively low search times when answering the questions (1.24 times per question on average).

\subsection{Why is StrategyQA hard?}
\label{sec:app_manual_analysis}
We manually analyze questions from StrategyQA and show some examples in Table \ref{tab:strategyqa_hard_examples} that could possibly explain this curious behavior. We use the question decompositions provided in the StrategyQA dataset as part of this analysis.
The key issues relate to difficulty of recognizing the implicit reasoning  using just the context, lack of intermediate quantities needed to do the required reasoning (e.g. number of children not deducible from context) and difficulty of handling abstractions and causal relationships. The reason why the StrategyQA model tends to rely on the distractors remains unknown and we leave it for future work.

\paragraph{Challenging inference from the compositional question to the decomposed question}  StrategyQA is designed such that the model should learn to decompose the question then retrieve evidence for each decomposed question. In row 1 of Table \ref{tab:strategyqa_hard_examples}, the original question is converted to comparing the number of descendants of Julius Caesar and Genghis Khan. This involves the implicit knowledge that \emph{if A has more children than B then probably more people today are related to A than B}. Note that this implicit knowledge is neither obvious nor definite.\footnote{If A was born much later than B, B could have more descendants today even though B had fewer children.} Acquiring and applying such knowledge can be challenging for the investigated methods. 

\paragraph{Decomposed questions requiring counting and listing}
The answer to ``How many kids did Julius Caesar have?'' (Row 1 Table \ref{tab:strategyqa_hard_examples}) is not directly shown in the gold paragraph. Instead, mentions of the children of Julius Caesar are scattered in the gold paragraph, and the model needs to collect those mentions across the document and count their number. Similarly, to answer ``How many commas are needed for a billion?'' the model needs to actually count the commas in the text.

\paragraph{Decomposed questions requiring abstraction}
The model needs to come up with the answer ``bad'' and ``good'' for the questions ``How sensitive is a grey seals hearing'' and ``How sensitive is a dog's hearing'' (Row 2 Table \ref{tab:strategyqa_hard_examples}).Coming up with such answers requires abstraction, and such abstraction is question dependent: the model needs to make certain abstractions that are helpful to answer the compositional question. 

\paragraph{Questions requiring the understanding of causal events} 
As shown in Row 4 of Table \ref{tab:strategyqa_hard_examples}, answering the question requires the model to reconstruct the causal chain ``MAO reduces serotonin'' + ``low level of serotonin causes depression'' and further infer that ``MAO can cause depression''. Note the mining of such causal events from text is itself far from being solved \cite{hahn2016before}.

\section{Conclusion}
In this work we investigate the impact of evidence quality on question answering on two open-domain QA datasets, StrategyQA and HotpotQA. Results show that (1) StrategyQA is less sensitive to the evidence quality than HotpotQA and (2) Boolean questions are less sensitive to the evidence quality than extractive questions. Further study shows this is mainly because the model does not yield a sufficiently high score on those examples with perfect evidence retrieval. Finally, we recommend that for the implicit decomposition  open-domain question answering problems with Boolean questions such as StrategyQA, researchers start by improving QA performance given the gold paragraphs rather than improving evidence quality. 


\bibliography{ref}
\bibliographystyle{acl_natbib}

\newpage
\appendix

\begin{table*}
\small
\begin{tabular}{p{1.5cm}|p{4.5cm} p{8.5cm}} \hline
 \textbf{Dataset} & \textbf{Question} & \textbf{Supporting Documents and Facts} \\ \hline \hline
  StrategyQA & Are Christmas trees dissimilar to deciduous trees? 
            & Doc 1: (\textbf{Christmas tree}): \textcolor{blue}{A Christmas tree is a decorated tree, usually an evergreen conifer, such as a spruce, pine or fir,} or an artificial tree  ... \\ 
        &   & Doc 2: (\textbf{Deciduous}): ... the term deciduous means ... \textcolor{red}{The antonym of deciduous in the botanical sense is evergreen.} \\ \hline
 HotpotQA  & The Oberoi family is part of a hotel company that has a head office in what city?  
          & Doc 1: (\textbf{Oberoi family}): \textcolor{blue}{The Oberoi family is an Indian family that is famous for its involvement in hotels, namely through The Oberoi Group.} \\
&         & Doc 2: (\textbf{The Oberoi Group}): \textcolor{red}{The Oberoi Group is a hotel company with its head office in Delhi.} Founded in 1934, the company owns and/or operates 30+ luxury hotels and two river ... \\ \hline
\end{tabular}
\caption{Examples from StrategyQA and HotpotQA. Only one annotator's evidence is shown for StrategyQA in the example. }
\label{tab:examples_hotpotqa_strategyqa}
\end{table*}

\section{Poisoned Context Generation}
\label{sec:appendix_poisoning}
\subsection{StrategyQA}
\label{sec:appendix_poisoning_strategyqa}

\paragraph{Sampling using Annotations}
Each question in StrategyQA has \textbf{3} gold paragraph annotators. We refer the $j^{th}$ supporting paragraph from annotator $i$ as $P_{i,j}$,  where $i \in \{1, 2, 3\}$ and $j \in \{1, ..., k_i\}$ (each annotator $i$ annotates $k_i$ gold paragraphs). When constructing the context $C$ for each example, we first randomly select one annotator $i$ from the three annotators, then randomly drop each paragraph $P_{i,j}$ from $k_i$ gold supporting paragraphs with the probability $p$. The quality of the context can thus be controlled by setting $p$. For example, setting $p=0.4$ would approximately result in context paragraphs with a recall of $0.4$ (on average).

\paragraph{Distractor Selection}
A pipeline is used to get the candidate distractors for the poisoning experiment for StrategyQA. We developed this pipeline to retrieve relevant paragraphs for StrategyQA but did not see any gains despite improvement in recall. We use the same pipeline now to generate distractors and ensure that none of the gold paragraphs are selected.

We first train a T5-large \cite{raffel2020exploring} to learn the gold decomposed questions that are provided by the annotators, then use each decomposed question as the query for first-stage retrieval. Then a BERT-based reranker is trained to rerank the initially retrieved documents. 

\paragraph{Training of T5 for decomposed questions} A T5-large is used to learn the gold decompositions given the compositional question. For example, given the compositional question ``Are more people today related to Genghis Khan than Julius Caesar?'', the model is trained to generate the gold decompositions ``How many kids did Julius Caesar have? How many kids did Genghis Khan have? Is $\#$2 greater than $\#$1?''. The model is trained for 14 epochs and the best model is selected by the best SARI score \cite{xu2016optimizing} on the validation set. 

\paragraph{First-stage retrieval given the generated decompositions} We run the trained T5 decomposer on each compositional question $q_i$ (for both train and validation set) and get the generated decompositions $D_i$ = [$d_{i, 1}$, ..., $d_{i, k}$] for each question (i.e., obtaining $k$ decomposed questions for question $i$). Then the decompositions in $D_i$ are used as the queries and BM25 \cite{robertson2009probabilistic} is used for first stage retrieval. We retrieve $400/k$ documents for each decomposed question. 

\paragraph{Reranking} We use a BERT-based reranker that is fine-tuned on MS-MARCO \footnote{\url{https://huggingface.co/amberoad/bert-multilingual-passage-reranking-msmarco}}. Then we further fine-tune it on StrategyQA to recognize the relevant paragraphs. The positive examples are constructed by concatenating each decomposed question with one gold paragraph. The negative examples mixed by (1) concatenating each decomposed question with a non-gold paragraph but with high BM25 score and (2) concatenating each decomposed question with a gold paragraph of another question. After applying this reranker to the first-stage retrieved paragraphs, the recall$@$10 for the training and validation set reaches 0.54 an 0.45. We use the reranked paragraphs as the distractors. 

\subsection{HotpotQA}
\label{sec:appendix_poisoning_hpqa}
We use a similar protocol to construct the poisoned context as StrategyQA. We randomly replace gold paragraphs with a probability $p$ and add the top ranked distractors using the paragraph scores from SemanticMRS~\cite{nie-etal-2019-revealing}, a strong re-ranker on HotpotQA. \footnote{We do not use the distractors provided by the HotpotQA dataset since many distractors have been reported to not be hard enough \cite{min-etal-2019-compositional}.}

The context is restricted to $k=6$ paragraphs to ensure a similar length distribution to that of StrategyQA. For the same recall value, setting $k=4$ would reduce the extent of distractor content and further improve the QA model scores. As we show in the experiment, the HotpotQA dataset at $k$=6 is already much easier than StrategyQA at $k$=4; hence we don't consider the easier setting of $k=4$.

The resulting poisoned data are referred to as HPQ-boolean (with all instances from the Boolean subset, 1855/209/458 for train/dev/test) and HPQ-extractive (with all instances from the extractive subset, 1855/209/6947 for train/dev/test). 

\section{UnifiedQA Model Details}
\label{app:uqa_details}
The UnifiedQA model is first fine-tuned on BoolQ and then on StrategyQA. The length of the input is limited to 1024 tokens.\footnote{Roughly $5\%$ to $10\%$ of questions in both poisoned datasets have a context exceeding 1024 tokens.} We use AdamW as the optimizer with a learning rate of 1e-4. The effective batch size is set to 32 (batch size of 2 with gradient accumulation set to 16).\footnote{Our model obtains a 0.725 test accuracy (averaged across 3 seeds) when trained on the gold paragraphs. This is comparable to the 0.723 accuracy of the RoBERTa model of \citet{geva-etal-2021-aristotle} reaches in the same setting.} 

\section{Ablating Context by Subsets}
\label{sec:app_ablating_context_all_subsets}
\begin{figure*}
     \centering
     \begin{subfigure}[b]{0.24\textwidth}
         \centering
         \includegraphics[width=\textwidth]{figures/contrastive_r3/easy_q+c_q.png}
         \caption{Acc. ``q+c'' vs ``q''}
         \label{fig:easy_qc_q_app}
     \end{subfigure}
     \begin{subfigure}[b]{0.24\textwidth}
         \centering
         \includegraphics[width=\textwidth]{figures/contrastive_r3/easy_q+c_q+gd.png}
         \caption{Acc. ``q+c'' vs ``q+gd''}
         \label{fig:easy_qc_qgd_app}
     \end{subfigure}
     \begin{subfigure}[b]{0.24\textwidth}
         \centering
         \includegraphics[width=\textwidth]{figures/contrastive_r3/easy_q+c_q+ds.png}
         \caption{Acc. ``q+c'' vs ``q+ds''}
         \label{fig:easy_qc_qds_app}
     \end{subfigure}
     \begin{subfigure}[b]{0.24\textwidth}
         \centering
         \includegraphics[width=\textwidth]{figures/contrastive_r3/easy_q+ds_q.png}
         \caption{Acc. ``q'' vs ``q+ds''}
         \label{fig:easy_q_qds_app}
     \end{subfigure}
     
     \begin{subfigure}[b]{0.24\textwidth}
         \centering
         \includegraphics[width=\textwidth]{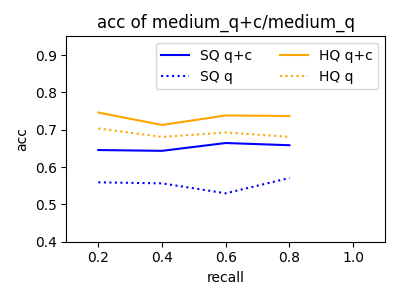}
         \caption{Acc. ``q+c'' vs ``q''}
         \label{fig:medium_qc_q_app}
     \end{subfigure}
     \begin{subfigure}[b]{0.24\textwidth}
         \centering
         \includegraphics[width=\textwidth]{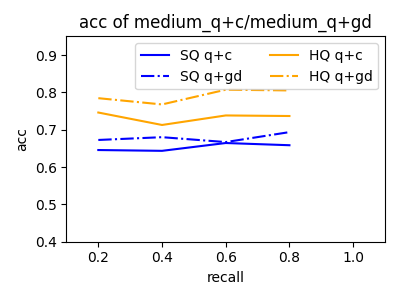}
         \caption{Acc. ``q+c'' vs ``q+gd''}
         \label{fig:medium_qc_qgd_app}
     \end{subfigure}
     \begin{subfigure}[b]{0.24\textwidth}
         \centering
         \includegraphics[width=\textwidth]{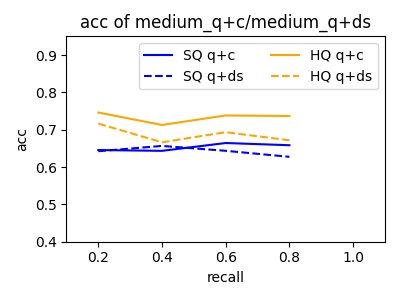}
         \caption{Acc. ``q+c'' vs ``q+ds''}
         \label{fig:medium_qc_qds_app}
     \end{subfigure}
     \begin{subfigure}[b]{0.24\textwidth}
         \centering
         \includegraphics[width=\textwidth]{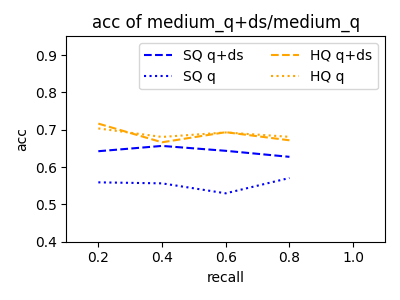}
         \caption{Acc. ``q'' vs ``q+ds''}
         \label{fig:medium_q_qds_app}
     \end{subfigure}
     
     \begin{subfigure}[b]{0.24\textwidth}
         \centering
         \includegraphics[width=\textwidth]{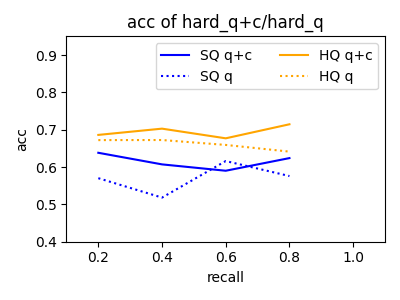}
         \caption{Acc. ``q+c'' vs ``q''}
         \label{fig:hard_qc_q_app}
     \end{subfigure}
     \begin{subfigure}[b]{0.24\textwidth}
         \centering
         \includegraphics[width=\textwidth]{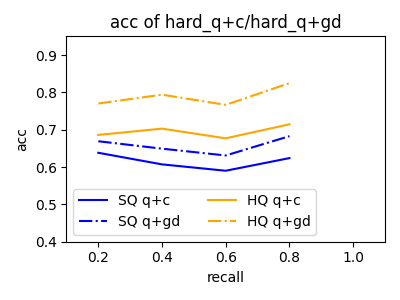}
         \caption{Acc. ``q+c'' vs ``q+gd''}
         \label{fig:hard_qc_qgd_app}
     \end{subfigure}
     \begin{subfigure}[b]{0.24\textwidth}
         \centering
         \includegraphics[width=\textwidth]{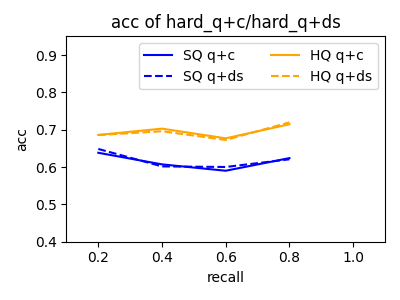}
         \caption{Acc. ``q+c'' vs ``q+ds''}
         \label{fig:hard_qc_qds_app}
     \end{subfigure}
     \begin{subfigure}[b]{0.24\textwidth}
         \centering
         \includegraphics[width=\textwidth]{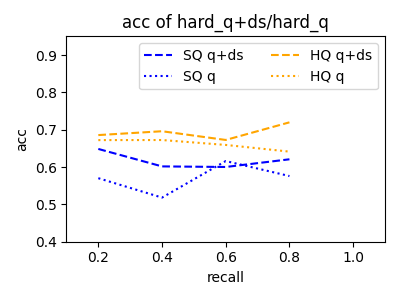}
         \caption{Acc. ``q'' vs ``q+ds''}
         \label{fig:hard_q_qds_app}
     \end{subfigure}
     
     \begin{subfigure}[b]{0.24\textwidth}
         \centering
         \includegraphics[width=\textwidth]{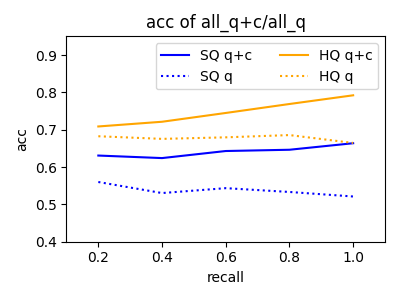}
         \caption{Acc. ``q+c'' vs ``q''}
         \label{fig:all_qc_q_app}
     \end{subfigure}
     \begin{subfigure}[b]{0.24\textwidth}
         \centering
         \includegraphics[width=\textwidth]{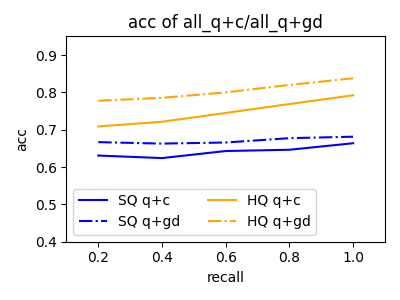}
         \caption{Acc. ``q+c'' vs ``q+gd''}
         \label{fig:all_qc_qgd_app}
     \end{subfigure}
     \begin{subfigure}[b]{0.24\textwidth}
         \centering
         \includegraphics[width=\textwidth]{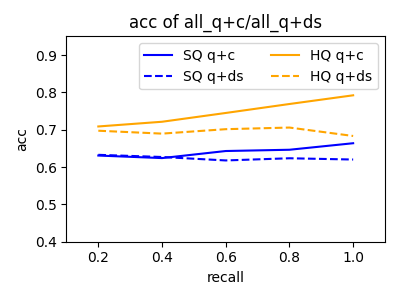}
         \caption{Acc. ``q+c'' vs ``q+ds''}
         \label{fig:all_qc_qds_app}
     \end{subfigure}
     \begin{subfigure}[b]{0.24\textwidth}
         \centering
         \includegraphics[width=\textwidth]{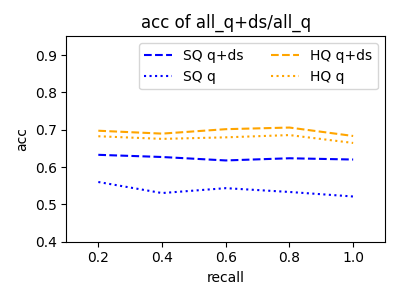}
         \caption{Acc. ``q'' vs ``q+ds''}
         \label{fig:all_q_qds_app}
     \end{subfigure}
\caption{Contrastive evaluation on all subsets of HotpotQA-boolean and StrategyQA. \cref{fig:easy_qc_q_app} to \cref{fig:easy_q_qds_app} are on the easy subset; \cref{fig:medium_qc_q_app} to \cref{fig:medium_q_qds_app} are on the medium subset; \cref{fig:hard_qc_q_app} to \cref{fig:hard_q_qds_app} are on the hard subset; \cref{fig:all_qc_q_app} to \cref{fig:all_q_qds_app} are on all subsets. }
\label{fig:sq_hq_contrastive_app}
\end{figure*}

This section shows the contrastive evaluation results for all subsets.
From the analysis of \cref{fig:sq_hq_contrastive_app} we can infer the HotpotQA model could distinguish good and bad context, whereas the StrategyQA model does not show a similar behavior. 

\paragraph{HotpotQA model relies more on the context when the context is in good quality, while relies more on the question itself when the context is in bad quality.} Note on the easy subset (\cref{fig:easy_qc_q_app}), the difference between ``q+c'' and ``q'' is very large for HotpotQA, indicating the model relies heavily on the context to predict the answers when the context is in good quality. However, the difference is much lower on the medium and hard subset (\cref{fig:medium_qc_q_app} and \cref{fig:hard_qc_q_app}), indicating that when the evidence quality is poor, the model tends to rely solely on the questions. 

In addition, when the evidence quality is bad, the HotpotQA model does not tend to rely on the distractors. As shown in \cref{fig:medium_q_qds_app} and \cref{fig:hard_q_qds_app}, the ``q+ds'' yields almost the same accuracy as ``q''. 

\paragraph{StrategyQA model relies heavily on the context regardless of the evidence quality. }
For all easy, medium and hard subsets, the difference between ``q+c'' and ``q'' remains relatively large for the StrategyQA model (\cref{fig:easy_qc_q_app}, \cref{fig:medium_qc_q_app} and \cref{fig:hard_qc_q_app}). This indicates the StrategyQA model relies heavily on the context to answer the questions regardless of the context quality. Moreover, the difference between ``q+c'' and ``q+ds'' is very small on all subsets (\cref{fig:easy_qc_qds_app}, \cref{fig:medium_qc_qds_app}, \cref{fig:hard_qc_qds_app}), and the ``q+ds'' number is generally much larger than ``q'' (\cref{fig:easy_q_qds_app}, \cref{fig:medium_q_qds_app}, \cref{fig:hard_q_qds_app}). These results combined indicate the StrategyQA model heavily relies on the distractors regardless of the evidence quality. 

\end{document}